\documentclass[10pt,conference]{IEEEtran}

\usepackage[ampersand]{easylist}

\usepackage{cite}
\usepackage{multirow}
\usepackage{amssymb}
\usepackage{verbatim}

\ifCLASSINFOpdf
   \usepackage[pdftex]{graphicx}
   \graphicspath{{figs/}}
   \DeclareGraphicsExtensions{.pdf,.jpeg,.png}
\else
   \usepackage[dvips]{graphicx}
   \graphicspath{{../figs/}}
   \DeclareGraphicsExtensions{.eps}
\fi
\usepackage[cmex10]{amsmath}
\usepackage{amsthm}

\usepackage{algorithmic}

\usepackage{array}

\ifCLASSOPTIONcompsoc
  \usepackage[caption=false,font=normalsize,labelfont=sf,textfont=sf]{subfig}
\else
  \usepackage[caption=false,font=footnotesize]{subfig}
\fi
\usepackage{url}

\begin{document}
%
\title{Training Deep Networks from Zero to Hero: avoiding pitfalls and going beyond}

\newif\iffinal
\finaltrue
\newcommand{\cmtid}{92}


\iffinal

\author{\IEEEauthorblockN{Moacir A. Ponti, Fernando P. dos Santos, Leo S. F. Ribeiro, Gabriel B. Cavallari}
\IEEEauthorblockA{ICMC -- Universidade de S\~ao Paulo (USP), S\~ao Carlos, SP, Brazil\\
Email: \url{ponti@usp.br}, \url{fernando_persan@alumni.usp.br}, \url{leo.sampaio.ferraz.ribeiro@usp.br}, \url{gabriel.cavallari@usp.br} } }


%

\else
  \author{Sibgrapi paper ID: \cmtid \\ }
\fi

\maketitle

\begin{abstract}
Training deep neural networks may be challenging in real world data. Using models as black-boxes, even with transfer learning, can result in poor generalization or inconclusive results when it comes to small datasets or specific applications. This tutorial covers the basic steps as well as more recent options to improve models, in particular, but not restricted to, supervised learning. It can be particularly useful in datasets that are not as well-prepared as those in challenges, and also under scarce annotation and/or small data. We describe basic procedures as data preparation, optimization and transfer learning, but also recent architectural choices such as use of transformer modules, alternative convolutional layers, activation functions, wide/depth, as well as training procedures including curriculum, contrastive and self-supervised learning. 
\end{abstract}


\IEEEpeerreviewmaketitle

\pagenumbering{roman}

\section{Introduction}
\label{secIntroduction}

Different fields were revolutionized in the last decade due to the huge investment in Deep Learning research. With the curation of large datasets and its availability, as well as popularization of graphical processing units, those methods became popular in all machine learning, pattern recognition, computer vision, natural language and signal/image processing communities~\cite{Bengio2013}. After becoming pervasive more broadly in related fields such as engineering, computer science and applied math~\cite{arpteg2018software, dimiduk2018perspectives, higham2019deep}, we observed a crescent number of projects and papers including deep learning techniques were adopted by practitioners from other fields in attempt to solve particular problems~\cite{esteva2019guide,christin2019applications,chalkidis2019deep}. Such hype raised concerns about the pitfalls in use of machine and deep learning methods. A remarkable example is the study of Roberts et al (2021) that, in a universe of over 2,000 papers using machine learning to detect and prognosticate for COVID-19 using medical imaging, found none of the models to be of potential clinical use due to methodological flaws and/or underlying biases~\cite{roberts2021common}.

In fact, training deep neural networks may be challenging in real world data. Using models as black-boxes, even with transfer learning -- a popular and widely used technique in this context --, can result in poor generalization or inconclusive results when it comes to small datasets or specific applications. In this paper, we focus on the main issues related to training deep networks, and describe recent methods and strategies to deal with different types of tasks and data. Basic definitions about machine learning, deep learning and deep neural networks are outside the scope of this paper. For those, please refer to the following as good starting points~\cite{goodfellow2016deep, Ponti2017everything, Bengio2013}.

\begin{figure}[hpbt]
\centering
\includegraphics[width=0.9\linewidth]{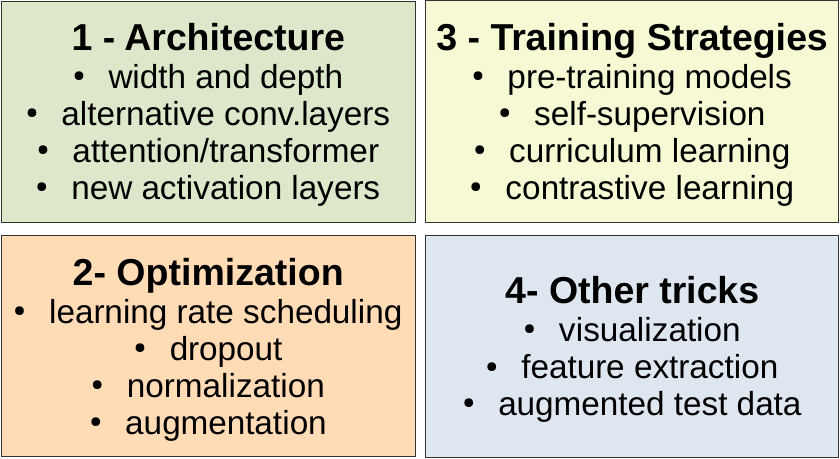}
\caption{Main concepts related to training of Deep Networks}
\label{fig:summary}
\end{figure}

We cover the basic steps to avoid common pitfalls as well as more recent options to improve models, in particular, but not restricted to, supervised learning in visual content. Those guidelines can be particularly useful in datasets that are not as well-prepared as those in challenges, and also under scarce annotation and/or small data. Figure~\ref{fig:summary} summarizes the main options to be considered. We describe importance of basic procedures but also recent architectural choices such as use of transformer modules, alternative convolutional layers, activation functions, wide and deep networks, as well as training procedures including as curriculum, contrastive and self-supervised learning\footnote{For an extended version containing details of the methods mentioned in thies paper see the arXiv extended version. For source-code related to this paper refer to:~\url{https://github.com/maponti/trainingdeepnetworks}.}.

\subsection{Notation}

Let $x \in X$ be examples from a training set containing $n$ instances, from which we may have target values or labels $y \in Y$. Such training set can be used to train a deep neural network (DNN) with multiple processing layers. For simplicity we define such neural network as a composition of functions $f_l(.)$ (related to some layer $l$) that has a set of parameters $\Theta_l$, takes as input a vector $x_l$ and outputs a vector $x_{l+1}$:
\begin{align*}
    f(x) = f_L \left( \cdots f_2(f_1(x_1, \Theta_1);\Theta_2) \cdots), \Theta_L \right)  
\end{align*}
$x_{1}$ is the input data coming from the training set, and functions $f_l(.)$ can represent different layers such as: Dense (or Fully Connected), Convolutional, Pooling, Dropout, Recurrent, among others. 
$\Theta_l$ represents all learnable parameters of a given layer. For example in dense layers those are matrices $W$ and bias values $b$, while in convolutional layers represent weights for convolutional kernels/filters. Also, let us have $z \in Z$ as examples from a test set used to evaluate the trained model. 

There are also non-sequential networks having different branches containing independent or shared parameters such as siamese or triplet networks but for which at least the output layer is shared among the branches. Other models operate using more than one independent networks: a remarkable example is the Generative Adversarial Network, which contains a discriminator and a generator function.

\section{How to start and common issues}
\label{secDrawback}

Common pitfalls and issues are due to overlooked details on the design of models. In this section we present a checklist, a kind of 7 Errors Game to begin with.

\subsection{Basic checklist (before trying anything else)}
\label{subChecklist}

\begin{figure}[hpbt]
\centering
\includegraphics[width=0.72\linewidth]{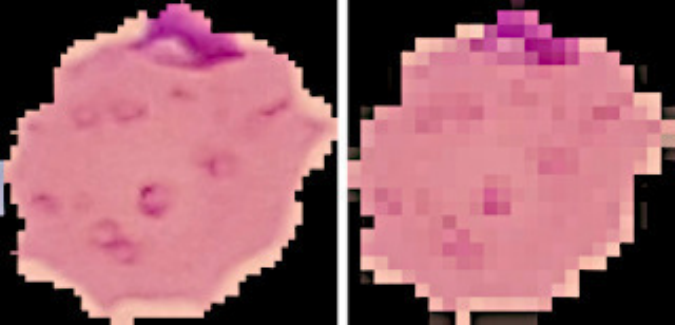}
\caption{Cell images resized to a size acceptable by a pre-trained network: left ($128\times128$) still retaining structures of the cell, right ($64\times64$) with insufficient details that would hamper learning.}
\label{fig:introImage}
\end{figure}

\noindent $\square$ \textit{1. Input representation is fair and target patterns are present in the data}. Make sure the input data is recognizible by a human or specialist, e.g., when undersampling and trimming an audio clip the expected patterns are still audible; when resizing images the objects to be recognized are still visible. For example in Figure~\ref{fig:introImage} we show two resized versions of an image to be used as input in a pre-trained neural network, however one of them clearly lost details of the cell that may be important for the task.

\vspace{3pt}

\begin{figure*}[hpbt]
\centering
\begin{tabular}{ccc}
\includegraphics[width=0.3\linewidth]{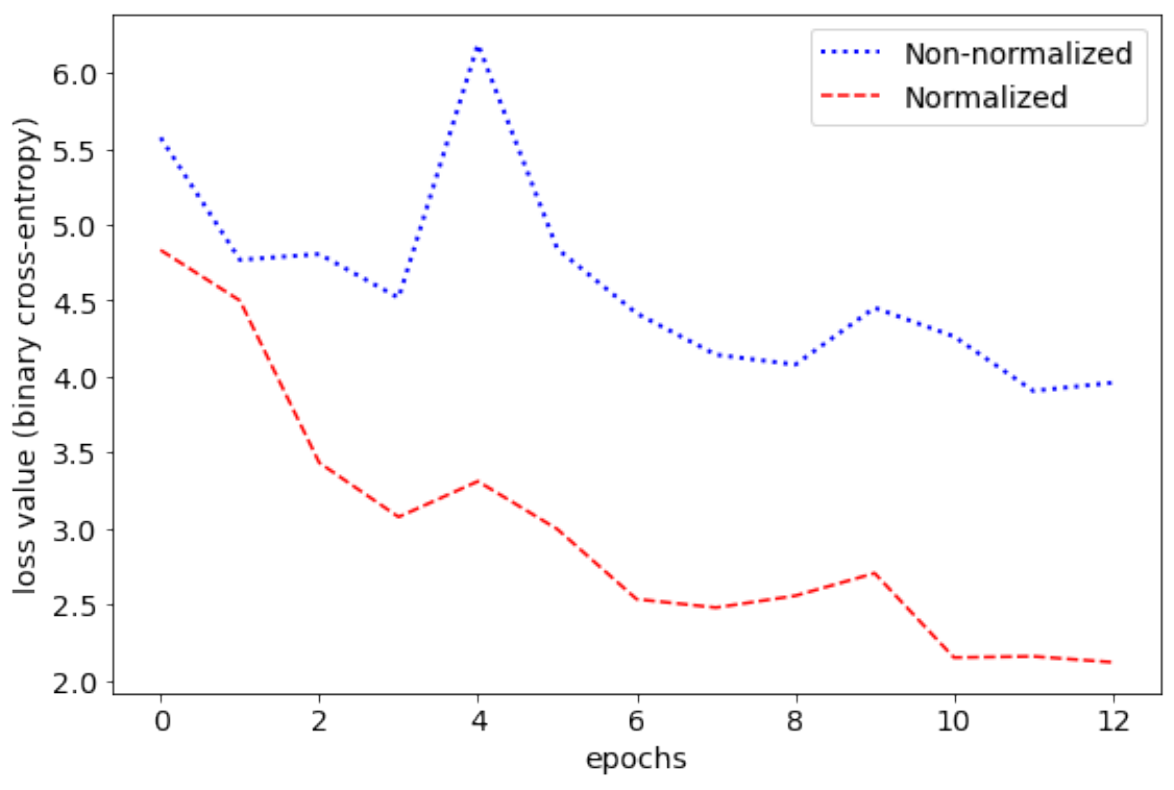}& \includegraphics[width=0.3\linewidth]{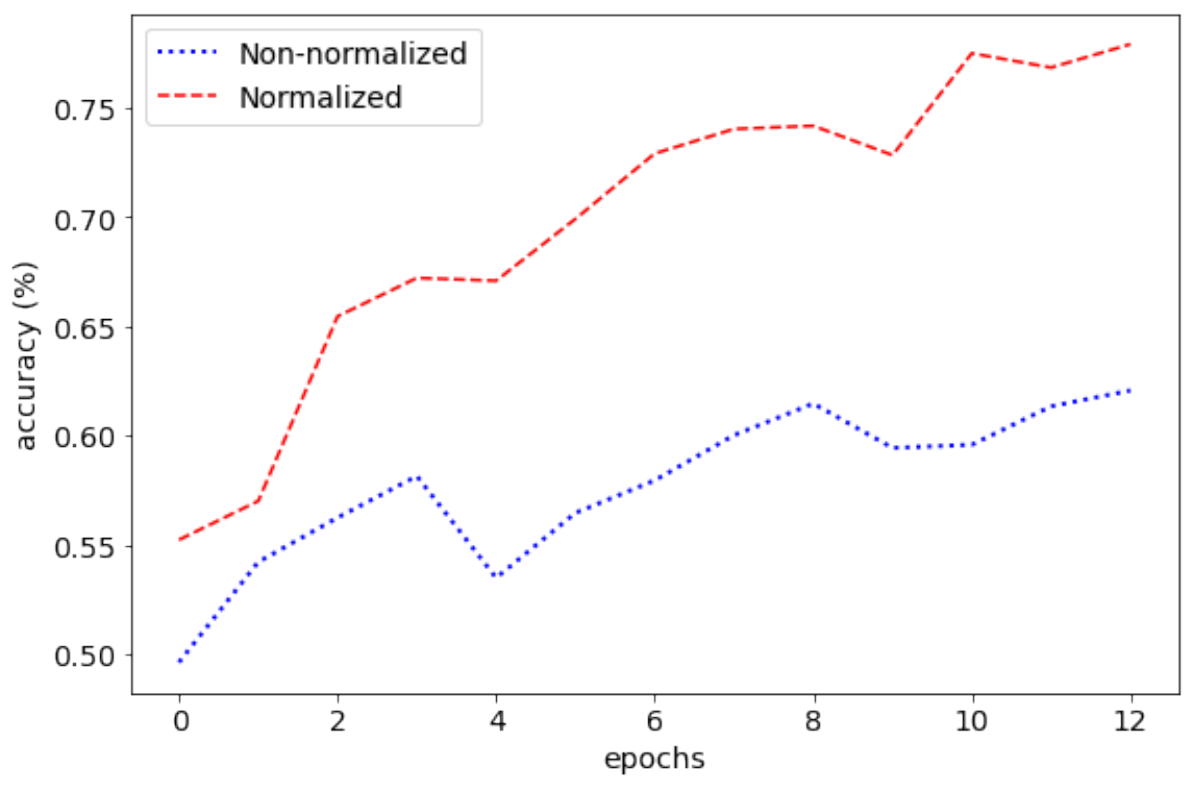} 
& 
\includegraphics[width=0.3\linewidth]{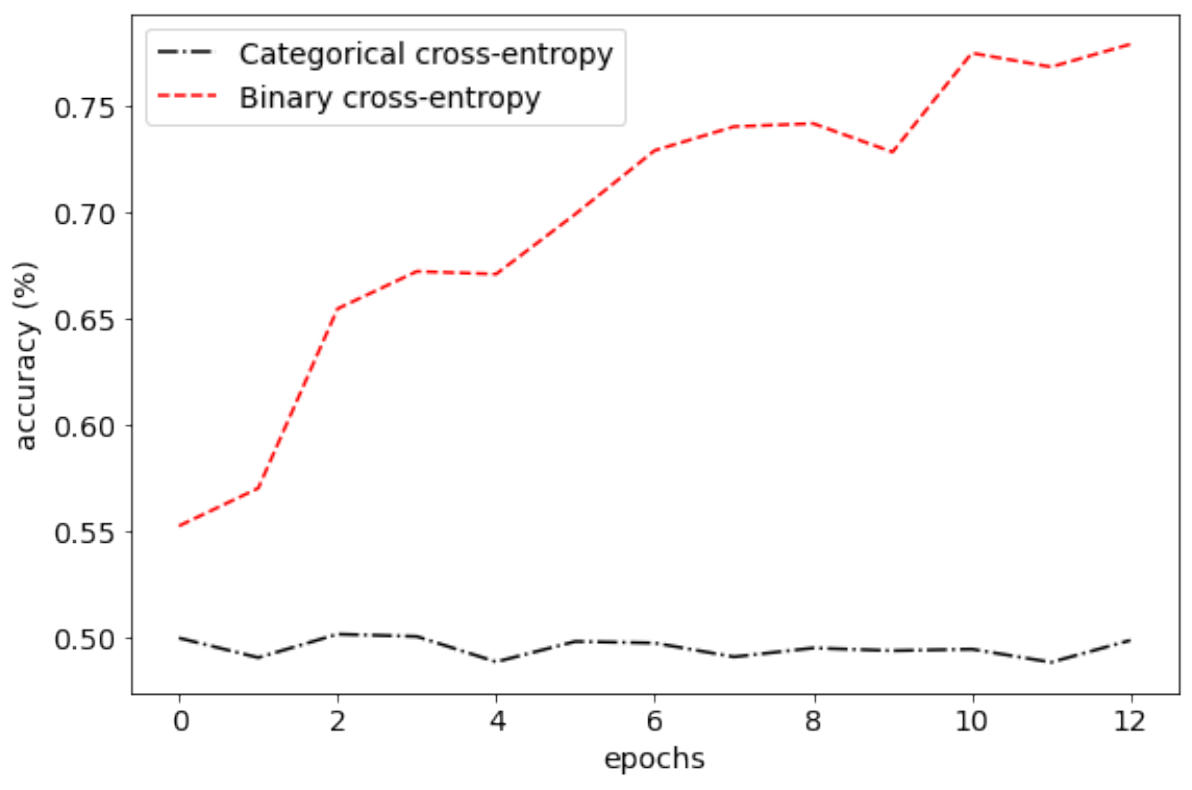}\\
    \small (a) Loss values for different normalization & \small (b) Accuracy for different normalization &  (c)  \small  Accuracy for different losses
\end{tabular}
\caption{Comparing loss curves (a) and accuracy (b) on the training set when training the same network with the same dataset for which the instances were normalized to 0-1 (dashed red line) and not normalized (dotted blue line), and the accuracy when using different loss functions (c).}
\label{fig:introLoss}
\end{figure*}

\noindent $\square$ \textit{2. Input data is normalized accordingly}.
DNNs do not work well with arbitrary ranges of numerical values. Common choices are 0-1 scaling (by computing and storing minimum and maximum values), or z-score standardization (by computing and storing mean and standard deviation). For example, in Figure~\ref{fig:introLoss}(a-b) we compared loss and accuracy curves using normalized and non-normalized versions of the training set.

\vspace{3pt}

\noindent $\square$ \textit{3. Data has quality (data-centric AI)}. After a decade of model frenzy, there has been a resurgence of concerns around data, that should be defined consistently, cover all important cases and be sized appropriately. Most datasets (even benchmark) have some wrong labels that hamper design of the model. In such case, recent work showed models with lower capacity (stronger bias) may be more resilient~\cite{northcutt2021pervasive}. Also there are datasets (in particular when built from different sources such as for Covid-19 detection in images) containing a significant amount of duplicates that should be removed.

\vspace{3pt}

\noindent $\square$ \textit{4. Both loss function and evaluation metrics makes sense}.\\
--- loss and evaluation must be adequate to the task and to the terms involved in its computation, e.g. in a multi-class classification task make sure you are comparing probabilities (vectors with unity sum). For regression tasks, error functions (such as mean squared error) are adequate, and for object detection the intersection over union (IoU)~\cite{rezatofighi2019generalized} is to be considered. Note the loss function must be differentiable, i.e., have a derivative! Metrics such as accuracy, area under the curve (AUC), Jaccard and cosine distances have particular interpretations and it is paramount to understand their meaning for the task you want to learn before using it;\\
--- check if the loss values are reasonable from the first to the last iteration, inspecting for issues such as overflow, e.g. the cross-entropy for 10 classes of a random classification result ($1/10$) should be no more than, approximately, $-\ln(0.1)=2.30$. Also, be sure your target (labels, range of values) matches what the network layer and its activation function outputs. For example, a sigmoid activation outputs values in the range $0-1$ for every neuron, while the softmax function outputs values so that the sum of all neurons is $1$. See Figure~\ref{fig:introLoss}(c) for the effects of using categorical vs binary cross entropy in a binary classification network in which the last layer contained only one neuron with sigmoid activation;\\
--- Plot the loss curve for the training and validation (whenever possible) loss values along iterations (or epochs). Loss value along iterations should decrease (fairly) smoothly and converge to near zero. If not, of when the training and validation curves are too different, investigate optimization details or rethink adequacy of the chosen model for the task.
    
\vspace{3pt}

\begin{figure}[hpbt]
\centering
\includegraphics[width=0.7\linewidth]{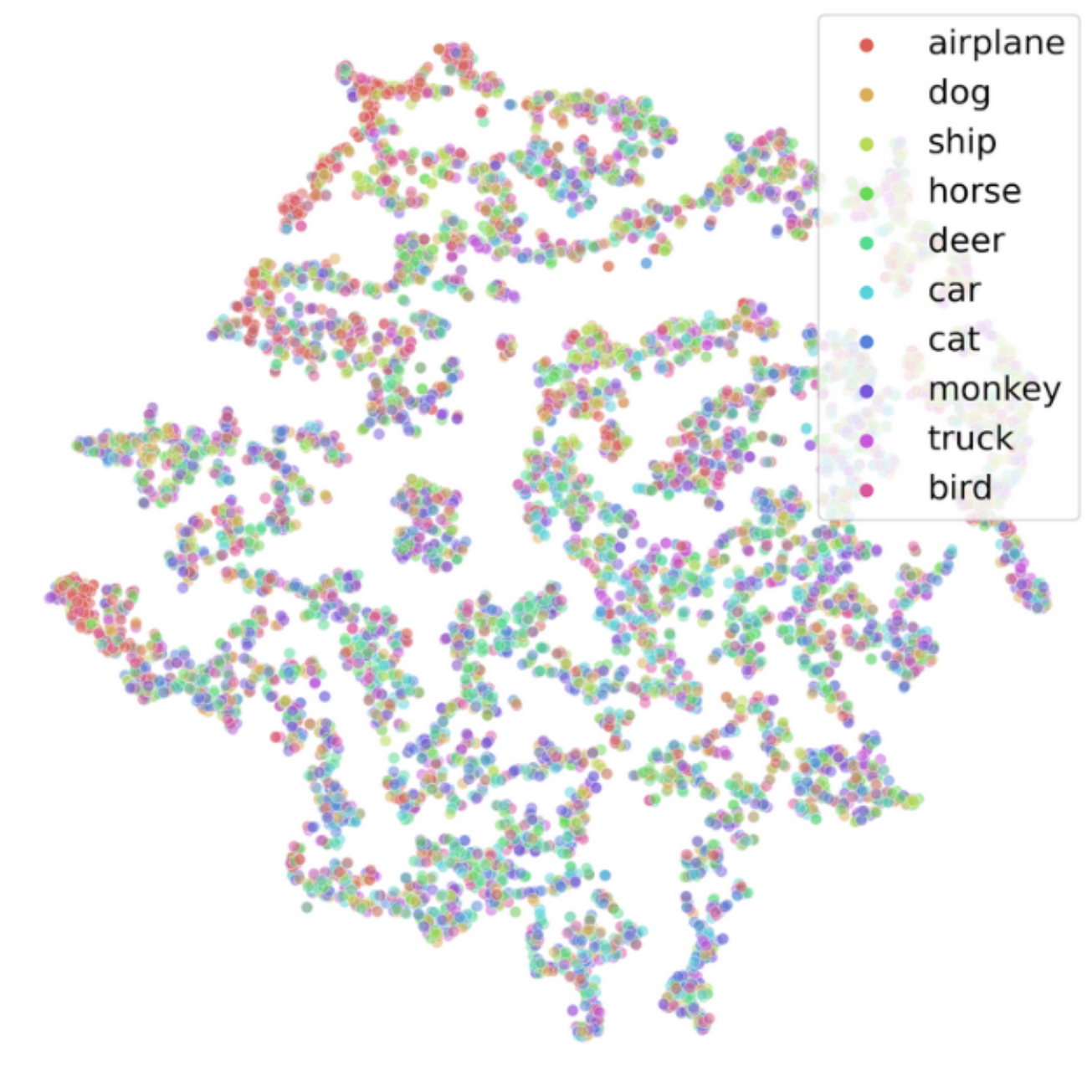}
\caption{t-SNE projection of the test set of STL-10 image features, extracted from the penultimate layer of a neural network that reached 35\% test accuracy, but for which the learned representations shows poor class separability.}
\label{fig:introProj}
\end{figure}

\noindent $\square$  \textit{5. Projected features has reasonable structure}. It is worth visualizing the learned feature space with tSNE~\cite{van2008visualizing} and UMAP~\cite{mcinnes2018umap} for example, by projecting into a 2d plane the learned features, e.g. the output of the penultimate layer (often the one just before the output/prediction layer). This complements the loss curve, and may show if such space makes sense in terms of the application, or if there was no actual convergence in terms of learning an useful representation as in the case of Figure~\ref{fig:introProj} in which a 10-class problem obtained a test accuracy of around 0.35, which is above random, but still far from having learning an useful representation as the same test set is projected and show no class structure.

\vspace{2pt}

\noindent $\square$ \textit{6. Model Tuning and Validation}. The correct way to adjust a model is to use a validation set, never the test set. If you have to make any decision regarding the data preparation, neural network design, training strategies, and other, such decisions have to consider only the training data available. In this scenario you may tune the model using metrics extracted for example via a $k$-fold cross validation on the training set. After all choices on network topology, training strategies, hyperparameters are made, then you evaluate the final model on the test set. Otherwise, the results (even for the test set) are biased and cannot be generalized.

\vspace{2pt}

\noindent$\square$ \textit{7. Use Internal and External Validation.} In particular for computed-aided diagnostics or deployment for decision-support, it is important to be extra careful with the data preparation, modeling and the conclusions. Methodological flaws and biases often lead to highly optimistic reported performance, but fail to be useful in practice. For example, a recent study identified 2,212 studies on COVID-19 diagnosis with chest radiographs and CT scans, from which 415 were screened, all having methodological flaws and/or underlying biases~\cite{roberts2021common}. We recommend reading and checking your study using PROBAST (tool to assess risk of bias and applicability of prediction model studies)~\cite{wolff2019probast} and/or CLAIM (checklist for artificial intelligence in medical imaging)~\cite{mongan2020checklist}, since they may be useful not only to health data but to assess models for other applications.

After you check-listed the items above, if results are still to be improved, we now have to set ourselves to investigate: (1) how difficult the learning task is, (2) what is the nature of the problem that makes it difficult and what options can be used to address it. Let us begin with difficult scenarios, as discussed in next sections.

\subsection{Small datasets and poor convergence}
\label{subSmall}

Learning under scarce data is known to be an issue with deep networks. For images, considering coarse-grained or category level data, i.e. the classes represent significantly different concepts such as in clothing and accessories Fashion-MNIST dataset, studies indicate a minimum of 1500 instances per class to allow learning. For fine-grained scenarios, i.e. the differences between concepts are more subtle, as in bird species CUB-200-2011 dataset (has around 60 instances per category), the problem may becomes harder if only the visual data is used in training. Therefore, if you have small data, consider transfer learning or feature extraction using DNNs (see Section~\ref{secTraining}), as well as architectures with less capacity (or reduced complexity). Data augmentation is also a possibility  (see Section~\ref{subAugmentation}), but if the original data is unrepresented, the augmented data will also be limited.

\subsection{Imbalance of target data in supervised tasks}
\label{subImabalance}

Ideally, in classification tasks, the number of examples available for each class should be similar, and for regression, training examples covering uniformly the whole range of the target data should be provided. When such supervision is not balanced with respect to the target data, one may be easily fooled by the loss function and evaluation metrics. Otherwise, one possible strategy is to weigh the classes so that the instances related to less frequent patterns will become more i the training process. Also, make sure you use metrics that evaluate how good the model along all the space of target values. In addition, data augmentation can be investigated as a way to mitigate for this imbalance (see Section~\ref{subAugmentation}).

\subsection{Complexity of models, overfitting and underfitting}
\label{subOverfitting}

Overfitting and underfitting represent undesired scenarios of learning and are related to the complexity of the models. Although it is not the scope of this paper to explain those phenomena (for a more complete explanation refer to~\cite{mello2018machine}), it is important to know how to diagnose them. 

\textit{Underfitting} usually occurs when the chosen architecture and training procedure are not well adapted to the task and/or the difficulty of the dataset at hand. The first symptom of this effect is a loss curve that converges to a value far from zero, or when there is no convergence at all.

\textit{Overfitting} is more common for deep neural nets since those are generally high capacity models, i.e. have a large number of trainable parameters that allow for a large space of admissible functions~\cite{mello2018machine}. It occurs when the network is excessively adjusted to the training set, approaching a model that memorizes the training set. Because DNNs often produce (near) zero error in the training set, it is harder to evaluate their generalization for future data. 

In an attempt to measure how deep networks may memorize the training set~\cite{Zhang2016understanding} uniformly randomized the labels of examples in benchmark datasets and showed that if the network has sufficient capacity, those are able to reach near zero loss (training error) by memorizing the entire training set. More recently~\cite{northcutt2021pervasive} showed lower capacity models may be practically more useful than higher ones in real-world datasets, which emphasizes the need for better data quality data and better evaluation, in particular external validation before finding a good balance between complexity of the model and its performance on a particular dataset.

\subsection{Attacks}
\label{subAttacks}

Deep networks learn features for a specific target task via a loss function that uses a specific training data. Because of its low interpretability, it is difficult to know which patterns from the input data were used to minimize the loss. For example, when counting white cells in blood smear images, if the purple color is present in all images with white, the optimization process has a huge incentive to use the purple color only as an indicator for white cells. Therefore, in future images, if there is purple dye in a blood smear medium (not the actual cells), the classifier may use this to incorrectly, but with a high confidence, classify the image as containing white cells. On the other hand, an image with white cells containing a different shade of purple may not be detected. The same can happen in soundscape ecology, for example when distinguishing from different bird species from its singing pattern. If there is a background noise, i.e. a critter, that usually sings at the same time of the day as some birds, the sound of the critter may be used by the network to detect the bird. In both scenarios, the features obtained after training are not the concept we wished to learn.

\begin{figure}[hpbt]
\centering
\includegraphics[width=0.75\linewidth]{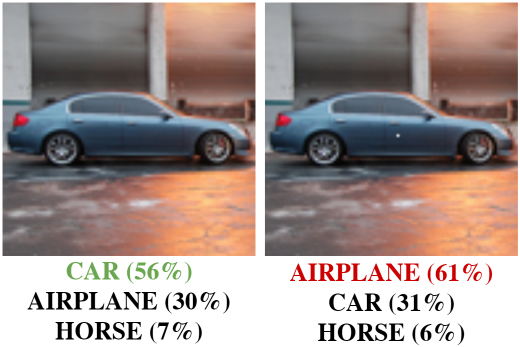}
\caption{Example of pixel attack in which the same network is given as input two testing images (not seeing during training state), the first without attack and the second with a one-pixel attack (see the white dot on the car's door), followed by the three most probable classes output by the network. In this image the pixel was included in a visible region to facilitate visualization.}
\label{fig:introAttack}
\end{figure}

For example, in Figure~\ref{fig:introAttack} we show two test images one without attack, and the other containing a visible one-pixel attack, in which images in the training set from a given class contain a white pixel in a fixed given position, biasing the model to use that white pixel in order to predict the class, while neglecting other visual concepts. In this case we deliberately included the pixel in a visible region, but one could include that in less obvious regions such as in the border, or even add subtle features, such as gradient with similar effects~\cite{nguyen2015deep}.

\section{Architecture options}
\label{secArchitecture}

\subsection{Types of convolution units}
\label{subConvolution}

Convolutions are of course the most important operation in CNNs, which means there are many studies in the literature bringing new ideas to this classic operation. 

Let the \textit{kernel size}, $k$, refers to the lateral size of each kernel in a convolutional layer, and we consider all those kernels to be square, so $(k,k)$ in size. Each kernel is applied one input channel to be accumulated for one output channel, which leads to each convolutional layer having the collection of $(c_{in}, c_{out})$ kernels for a total of $(k, k, c_{in}, c_{out})$ learnable weights. The \textit{stride} of a conv. layer refers to the step between each ``application'' of a kernel when it ``slides'' over the image, in the classic operation this step size is always one. 

\noindent $\square$ \textit{$1\times 1$ convolution} can be useful for reducing computations further into networks by combining values along the channels of a single pixel. 
These operations do not take into account any neighbourhood, but perform the role of weighing and collecting information for each pixel on all $c_{in}$ channels and outputting at a new channel dimensionality $c_{out}$. 


\noindent $\square$ \textit{Transposed Convolutions} play the role of a learnable, weighted upsampling operation in generative networks, autoencoders and pixel-to-pixel models (e.g. segmentation tasks). The concept simulates a \textit{fractional stride}, so before applying the kernels a feature map is padded with zeros between spatial dimensions. When using this operation it is important to choose $k$ as an even number to avoid the ``checkerboard effect'', per \cite{odena2016checkerboard} on the effect.

\noindent $\square$ \textit{Spatially Separable Convolutions} save on computation by breaking a larger convolution operation into two smaller operations. This is usually accomplished by making a convolution with $k\times k$ kernels into a $1\times k$ followed by a $k\times 1$ operation. 

\noindent $\square$ \textit{Depthwise Separable Convolutions} Follow a similar principle to spatially separable ones by also breaking the traditional operation into two more efficient ones. First, the feature map is convolved with $c_{in}$ $k\times k$ kernels, but instead of summing the resulting activations as usual, the $c_{in}$ matrices go through a $1\times 1$ convolution to map the output to have the desired $c_{out}$ number of feature maps. This yields the same output shape as the traditional operation, but at a fraction of the cost. 
\subsection{Width, Depth and Resolution}
\label{subDepth}
Techniques for designing deep networks have evolved considerably since AlexNet \cite{Krizhevsky2012AlexNet} won the 2012 ImageNet challenge. One of the main fronts of discussion is around scaling networks up or down find a balance between accuracy and memory/computational efficiency. Width, Depth and Resolution are strategies with different pros and cons.

\noindent $\square$ \textit{Wider Nets are easier to train} and are able to capture finer details in images (such as background information). Increasing width  increases computational cost exponentially~\cite{Zagoruyko2016WideResNet} 

\noindent $\square$ \textit{Deeper Nets perform better on ``well-behaved'' datasets}~\cite{Nguyen2021WideDeep}, such as single-object classes with ``clear'' objects, while wider nets did better on classes that represent scenes (e.g. ``bookshop'', ``seashore''). 

\noindent $\square$ \textit{Scaling Depth, Width and Resolution Together} yields the best results for a wide range of tasks and desired accuracies. Frameworks for scaling the three variables together were presented in EfficientNet \cite{Tan2019EfficientNetV2} and MobileNet \cite{Sandler2018MobileNetV2}.

\subsection{Pooling}
\label{subPooling}

Pooling layers have been a staple of CNNs since their introduction; These downsampling operations are useful both for saving on computation, memory, and for summarising feature maps as networks get deeper.

\noindent $\square$ \textit{Max Pooling} 
is the most widely used method for classification as it enforces discriminative features within the network.

\noindent $\square$ \textit{Average Pooling} was the first pooling approach and is currently used in Generative Adversarial Networks as in those models they better match the upsampling layers of generators.

\noindent $\square$ \textit{Strided Convolutions}, with step size $> 1$ are a way of implementing ``learnable pooling''. Less common in classification CNNs, more common in GAN designs.

\noindent $\square$ \textit{Blur Pooling} is a solution proposed by~\cite{zhang2019blurpooling}. Their findings showed that current operations break the shift equivariance expected of CNNs and proposes that pooling operations are first densely evaluated, blurred by a low pass filter and only then subsampled. This improves shift equivariance.

\subsection{Transformers and ViT}
\label{subTransformer}

Transformers are a recent architecture created primarily for language tasks \cite{Vaswani2017Transformer}. It relies on self-attention as the defining mechanism of its layers. 
Self-attention is very different from convolutions and from recurrent layers in that very little inductive bias is taken into account for its mechanism. 

On attention layers, the relevance of one item to all other items including itself is estimated so that each item becomes a weighted average of each most relevant counterpart. This is done by learning three projection matrices $W_q,W_k,W_v$ (similar to 3 dense layers) applied to the input items $X$:
\begin{equation}
    Z = \sigma\left(\dfrac{(XW_q)(XW_k)^T}{\sqrt{d_q}}\right)XW_v
\end{equation}
where $\sigma$ is the softmax function that makes the resulting \textit{attention weights} (the result of $(XW_q)(XW_k)^T$) behave as a probability function that weights the values $XW_v$. $d_q$ is the dimensionality of each item.

The architecture was quickly applied to compute vision tasks as well. Notable examples being the ViT \cite{dosovitskiy2021ViT} for image recognition, iGPT \cite{chen2020iGPT} for image generation. 
While ongoing work with this architecture is exciting, the lack of inductive bias means that those models require much more training data than CNNs. ViT for example cannot be trained from scratch on the ImageNet dataset alone and perform well.


\section{Improving optimization}
\label{secOptimization}

Optimization choices: algorithm, learning rate (or step size) and batch size, matters when using deep networks for learning representations. Using default options with arbitrary optimizers may lead to suboptimal results. Also, normalization, regularization and the sample size may significantly influence the optimization procedure.

\subsection{Optimizer and batch size}
\label{subOptimizer}

The original Gradient Descent algorithm computes the gradient at one iteration using all training data. Stochastic Gradient Descent (SGD) is an approximation that allows calculating the gradient of the cost function based on a random example or a small subset of examples (minibatch). The regular SGD is a conservative but fair choice, as long as the learning rate and batch size are well defined. In fact, nearly every state-of-the-art computer vision model was trained using SGD, for example ResNet~\cite{He2016ResNet}, SqueezeNet~\cite{iandola2016squeezenet}, Faster R-CNN~\cite{Shaoqing2015}, Single Shot Detectors~\cite{WeiSSD2015}.

Adaptive methods such as Adam and RAdam are good alternatives, requiring smaller learning rate (LR) values (0.001 or lower) and larger batch sizes when compared to SGD, which is less sensitive to batch size choice and LR choice is often around 0.01. Momentum can be used as an to accelerate convergence of regular SGD, however it adds another hyperparameter (the velocity weight) to be set.

\subsection{Learning rate scheduling }
\label{subLearning}

A bad learning rate choice may ruin all other choices. Because the parameter adjustment is not uniform along the training process, a learning rate/step adaptation using scheduling should always be considered:\\
\noindent $\square$ \textit{Step Decay}, decreases the learning rate by some factor along the epochs or iterations, e.g. halving the value every 10 epochs, \\
\noindent $\square$ \textit{Exponential Decay}, reduces the learning rate exponentially.\\
\noindent $\square$ \textit{Cosine Annealing}, continuously decreases step to a minimum value and increase it again, which acts like a simulated restart of the learning process \cite{loshchilov2016sgdr}. 

\subsection{Normalization}
\label{subNormalization}

Normalizing data is a staple of classic machine learning. Since deep models are composite functions, it is beneficial to keep intermediary feature maps within some range:

\noindent $\square$ \textit{Batch Norm.} (BN), a widely known technique, it was introduced by \cite{Ioffe2015BatchNorm} to accelerate training of deep networks; it works like a layer that standardizes the feature maps across each input in a minibatch (hence the name). As learning progresses it also learns an average mean and standard deviation across the dataset that can then be used for doing single sample inference. Santurkar et. al. \cite{Santurkar2018BatchNormFix} showed that BN's advantage comes from making the optimization space smoother.

\noindent $\square$ \textit{Instance Norm.} can be also designed as a layer, but instead of performing standardization across input samples, it does so for each channel of each individual sample. It's performance is worse than BN for recognition. It was designed specifically for generative and style transfer models~\cite{Ulyanov2016InstanceNorm}.

\noindent $\square$ \textit{Layer Norm.}  Performs standardization for each individual sample but takes mean and standard deviation from all feature maps. It was created \cite{Ba2016LayerNorm} because BN cannot be applied in recurrent networks since the concept of a batch is harder to define in that context. Layer Norm. is also used on most Transformer Implementations.

\subsection{Regularization via Dropout}
\label{subRegularization}

Comprises mechanisms to help find the best parameters while minimizing the loss function. During the convergence process of deep networks, several combinations of parameters $\Theta$ can be found to correctly classify the training examples. Hence, \textbf{Dropout}\cite{hinton2012improving} works by deactivating $p\%$ of neurons, mainly after dense layers. This avoids some neurons to over-specialize/memorize specific data. At each iteration of training, dropout provides different subsamples of activations, i.e different stages of the network. Consequently, this mechanism prevents overfitting during training. During inference dropout is turned off so that all neurons are activated.




\subsection{Data Augmentation}
\label{subAugmentation}

Unlike the other optimization techniques mentioned before, which work to improve performances by acting on the network structure, data augmentation techniques focus exclusively on increasing the size and variability of the training set~\cite{shorten2019survey}. Conceptually, it generates new instances derived from the original training set by manipulating the features and replicating the original label to the generated example. Thus the training set becomes more variable and larger. Data augmentation can also be used to balance datasets (see Subsection~\ref{subImabalance}), controlling one of the drawbacks of deep learning~\cite{leevy2018survey}. A recurrent concern in these techniques is to ensure that the transformation performed does not alter its concept.

\section{Training procedures beyond the basics}
\label{secTraining}

The regular approach for training deep networks is to design its topology, define its training strategies, randomly initializing all parameters and then \textit{train from scratch}. However such networks are both data-hungry and highly sensitive to initialization. To overcome those issues, weight warmup procedures were studied, such as first training an unsupervised autoencoder~\cite{cavallari2018unsupervised} and then use its encoder weights as initialization. In addition, a widespread approach is to download models pre-trained using a large datasets such as ImageNet in the case of image classification~\cite{kornblith2019better}. This is called transfer learning, and assumes the model has generalization capability. Due to the hierarchical structure of deep networks, in which different layers provide different levels of attributes, even different image domains may benefit from pre-training~\cite{ponti2019supervised, dos2019generalization}.

\noindent-- \textbf{Transfer learning from pre-trained weights}:
\begin{enumerate}
    \item remove the original output layer, design a new output layer and randomly initialize its weights;
    \item freeze the remaining layers, i.e. making the layers not trainable, by not allowing their parameters to be updated;
    \item train the last layer for a number of epochs.
\end{enumerate}

\noindent-- \textbf{Fine-tuning} after transfer learning, unfreeze and train a subset of layers using a small learning rate (often $10^{-4}$ or even less). As a rule of thumb, one starts by unfreezing the layers closer to the top (output) of the network and, the more data one has, more layers can made trainable. Use with care: if your dataset is small, beware not to overtrain.

\noindent-- \textbf{Pre-trained nets can be used as feature extractor} in scenarios with small sets of data, in which even transfer learning would be unfeasible. For this, perform a forward pass and get the activation maps of a given layer as a feature vector for the input data. Getting the output from the penultimate layer is a fair choice since this represents input data globally~\cite{ponti2019supervised}. However, one can also insert a global pooling layer just after a convolutional layer to summarize the data. Previous studies show that combinations coming from different layers improve the representation~\cite{dos2019alignment, zheng2019cnn}. 

Alternatively to the use of a global pooling layer, get all activation maps/values (often high dimensional) and carry out a separate dimensionality reduction, for instance using Principal Component Analysis (PCA). With the extracted features one can proceed with external methods such as classification, clustering, and even anomaly detection~\cite{dos2019generalization}.

In the next sections we will cover training strategies beyond the transfer learning approach.

\subsection{Curriculum Learning}
\label{subCurriculum}

\begin{figure}[!ht]
\centering
\includegraphics[width=3.7in]{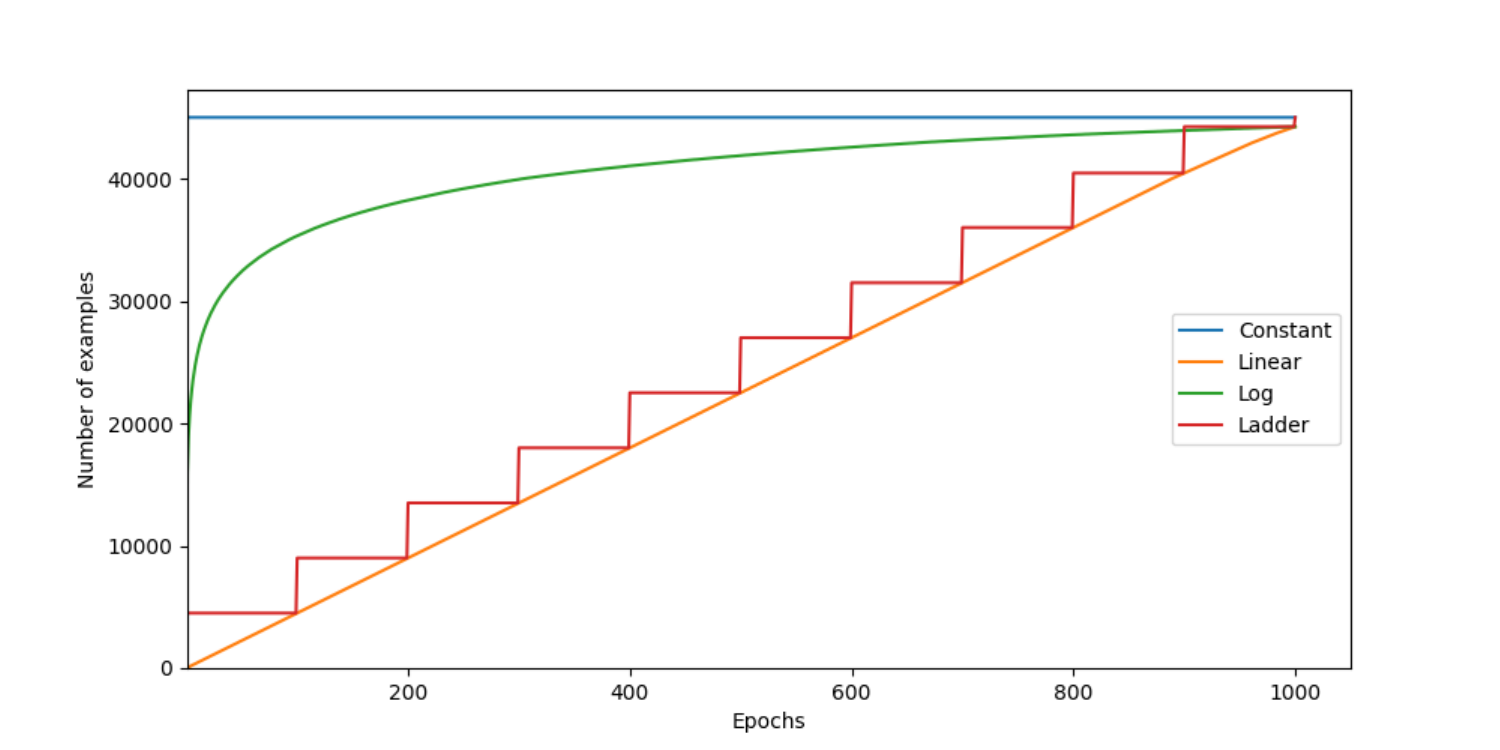}\\
\caption{Pacing functions: different increment functions can be chose to increase the variability of data in the training set. In this example, we have constant (conventional training), linear, logarithm, and ladder functions.}
\label{fig:curriculumPacing}
\end{figure}

This concept is based on the human strategy of creating a study script, in which a teacher elaborates a student's learning order, facilitating training~\cite{bengio2009curriculum,hacohen2019power}. With the premise that part of the data (or the task) at hand is easier than others to be learned, instead of trying to train all model at the same time with randomly sampled data, it is possible to define an order of instances or tasks. The basic technique works with instances by defining: a scoring function and a pacing function. The scoring function is a metric to sort the training examples from the easiest to the most difficult, e.g. a shallow classifier confidence. The pacing function, e.g. linear, exponential (see Fig.~\ref{fig:curriculumPacing}), dictates when to incorporate more examples into the training set. Note that unbalanced scenarios can be harder when applying curriculum learning. Also, learning rate have to be properly investigated so that not to degrade performance~\cite{hacohen2019power}. Curriculum learning can also be applied as a sequence of tasks, where the easiest task is performed before the most difficult ones~\cite{Bui2018Sketching}.

\subsection{Contrastive/Distance Learning}
\label{subContrastive}


Deep learning is often used to learn representations using tasks such as classification, e.g. via CNNs, and reconstruction, e.g. via Autoencoders. Since good semantically aware representations are of interest to more tasks than classification/reconstruction, there are many ways of exploring deep learning. An alternative way is using \textit{contrastive learning}, which consists of using losses designed around the task of learning (dis)similarities between instances. 


This idea was first introduced in \cite{hadsell2006} as the contrastive loss, where two instances of a network compute representations from two samples and if semantically distinct the loss is the similarity between those representations, if semantically similar the loss is the distance. 


Variants of the original contrastive loss followed, such as triplet loss which works using triplets of instances. While initial applications included face recognition \cite{schroff2015Facenet} and content-based image retrieval \cite{Bui2018Sketching}. More recently, self-supervised learning made use of this strategy.

\subsection{Self-supervised Learning}
\label{subSelf}

Given a task and sufficient labels, supervised learning can solve it. But large amounts of manually labeled data are often costly and time-consuming to obtain. Sometimes real-world applications require concepts that are outside the scope of standard datasets. And in some cases, vast amount of unlabelled images is readily available.

Self-supervision is a form of unsupervised learning where the data itself provides the supervision. It relies on pretext tasks that can be formulated using only input data. For example one can produce surrogate (or pseudo) labels for classification or design systems that learn to compare, using a contrastive learning strategy. Those are than used to pre-train a model instead of relying only on large-scale benchmark datasets. Methods include predictions of data rotation, relative positions, maximization of mutual information, cluster-based discrimination and instance discrimination. Relevant works are SimCLR~\cite{chen2020simple}, SwAV~\cite{caron2020unsupervised} and Barlow Twins~\cite{zbontar2021barlow}. For a deeper understanding of those models and an extensive view of another methods, we recommend~\cite{jing2020self} and ~\cite{liu2021self}.

\section{Running the final mile to improve predictions}
\label{secPrediction}

After previous strategies were explored, some other tricks may produce small but valuable improvement.

\subsection{Activation Functions beyond ReLU}
\label{activationFunction}

The Rectified Linear Unit (ReLU) became the standard activation function for hidden layers of deep networks because it avoids saturation which can cause training to slow down due to near-zero gradients. The problem is that values of ReLU near zero produce non-useful or bad estimates for the gradient. Numerically, it may lead to neurons that stuck completely in the negative side and always outputs zero for the training set. Swish and Mish functions were proposed to improve this. 

Swish is a gated version of Sigmoid and defined as $s(x) = x\cdot \sigma(\beta x)$, where $\sigma(.)$ is the Sigmoid function and $\beta$ is a hyper-parameter that can be adjusted arbitrarily or trained. Mish is defined as $m(x) = x \cdot \tanh(\ln(1+e^x))$, which is bounded below and unbounded above and the range is approximately $[-0.31, \infty]$. Small but consistent improvement were observed when using Swish and Mish instead of ReLU in hidden layers of the network, with a slight advantage for Mish.

\subsection{Validation and Test-Augmented Data}
\label{subTest}

To assess the robustness of the trained model, one can compare its performance in the original validation set with a perturbed version containing only modified versions of the instances, i.e. by translation, noise injection, etc, one can decide to include those perturbed data in the training set.

To improve final performance after the model is trained, given a test example $x$, obtain augmented versions: $x^{(1)}, x^{(2)}, \ldots x^{(m)}$, and combine the network predictions $\hat{y}^{(1)}, \ldots \hat{y}^{(m)}$ (via average, majority voting or other~\cite{ponti2011combining}).

\section{Conclusion}
\label{secConclusion}

This paper offers as a reference to allow researchers and practitioners to avoid major issues and to improve their models with less usual techniques. While DNNs have high generalization capacity and allow significant transfer learning, there are important concepts that require attention to allow learning. The basic recommendations for machine learning should be observed, and given the representation learning nature of deep networks, employ other practices that mainly try to improve the representations, not only the main objective function. Going beyond the basic techniques, leveraging unlabeled data and carefully designing optimizations steps beyond the basics may be the way towards more reliable models.

\iffinal
 \section{Acknowledgments}
The authors are grateful to FAPESP grants  \#17/22366-8, \#19/07316-0 and \#19/02033-0 and CNPq 304266/2020-5.
\fi

\bibliographystyle{IEEEtran}
\bibliography{example}
\end{document}